\documentclass{article}

\usepackage{microtype}
\usepackage{graphicx}
\usepackage{subcaption}
\usepackage{booktabs} 

\usepackage{hyperref}

\usepackage{url}
\usepackage{graphicx}
\usepackage{wrapfig}

\usepackage{array, makecell}

\usepackage{xfrac}
\usepackage{csquotes}
\usepackage{enumitem}
\usepackage{hhline}
\usepackage{caption}
\usepackage{multirow}
\usepackage{multicol}
\usepackage{amssymb}
\usepackage{microtype}      
\usepackage{amsfonts}

\usepackage{chngcntr}

\usepackage[outercaption]{sidecap}
\usepackage{floatrow}
\usepackage{float}
\usepackage{multirow}
\usepackage{booktabs}
\usepackage{wrapfig}
\usepackage[bottom]{footmisc}

\usepackage{xcolor} 
\definecolor{LightGray}{gray}{0.97}

\usepackage[ruled]{algorithm2e}

\usepackage[accepted]{icml2023}

\usepackage{amsmath}
\usepackage{amssymb}
\usepackage{mathtools}
\usepackage{amsthm}

\usepackage{csquotes}

\usepackage[capitalize,noabbrev]{cleveref}

\usepackage{pifont}

-lmtk10
-lmtl10

\theoremstyle{plain}

\theoremstyle{definition}

\theoremstyle{remark}

\usepackage[textsize=tiny]{todonotes}

\usepackage{amsmath,amsfonts,amssymb, amsthm}
\usepackage{mathtools}

\usepackage{scalerel}
\usepackage{stackengine,wasysym}

\usepackage{MnSymbol}%
\usepackage{wasysym}%
\usepackage{multirow}

\usepackage{dsfont}

\renewcommand{\vec}[1]{\ensuremath{\mathbf{\MakeLowercase{{#1}}}}}

\def\gE{{\mathcal{E}}}

\def\gG{{\mathcal{G}}}

\def\gO{{\mathcal{O}}}
\def\gP{{\mathcal{P}}}

\def\gU{{\mathcal{U}}}

\def\sR{{\mathbb{R}}}

\newcommand{\xv}{\vec{x}}

\newcommand{\cv}{\vec{c}}

\newcommand{\Nt}{\mathrm{N}}
\newcommand{\Dt}{\mathrm{D}}
\newcommand{\Kt}{\mathrm{K}}

\newcommand{\Ft}{\mathrm{F}}
\newcommand{\Ht}{\mathrm{H}}

\newcommand{\ev}{\vec{e}}

\definecolor{mydarkblue}{rgb}{0,0.08,0.45}
\definecolor{mathematicablue}{rgb}{0.11, 0.25, 0.467}
\hypersetup{colorlinks,citecolor={mydarkblue},urlcolor={mydarkblue}, linkcolor={red}}

\icmltitlerunning{Learned Gridification for Efficient Point Cloud Processing}

\begin{document}

\setlength{\abovedisplayskip}{4pt}
\setlength{\belowdisplayskip}{4pt}

\twocolumn[
\icmltitle{Learned Gridification for Efficient Point Cloud Processing}


\icmlsetsymbol{equal}{*}

\begin{icmlauthorlist}
\icmlauthor{Putri A. van der Linden$^*$}{uva}
\icmlauthor{David W. Romero$^*$}{vu}
\icmlauthor{Erik J. Bekkers}{uva}
\end{icmlauthorlist}

\icmlaffiliation{vu}{Vrije Universiteit Amsterdam, The Netherlands}
\icmlaffiliation{uva}{University of Amsterdam.}

\icmlcorrespondingauthor{Putri A. van der Linden}{p.a.vanderlinden@uva.nl}
\icmlcorrespondingauthor{David W. Romero}{d.w.romeroguzman@vu.nl}

\icmlkeywords{Machine Learning, ICML}

\vskip 0.3in
]



\printAffiliationsAndNotice{\icmlEqualContribution} 

\begin{abstract}

Neural operations that rely on neighborhood information are much more expensive when deployed on point clouds than on grid data due to the irregular distances between points in a point cloud. In a grid, on the other hand, we can compute the kernel only once and reuse it for all query positions. As a result, operations that rely on neighborhood information scale much worse for point clouds than for grid data, specially for large inputs and large neighborhoods.\newline 
In this work, we address the scalability issue of point cloud methods by tackling its root cause: the irregularity of the data. We propose \textit{learnable gridification} as the first step in a point cloud processing pipeline to transform the point cloud into a compact, regular grid. Thanks to gridification, subsequent layers can use operations defined on regular grids, e.g., \texttt{Conv3D}, which scale much better than native point cloud methods. We then extend gridification to point cloud to point cloud tasks, e.g., segmentation, by adding a \textit{learnable de-gridification} step at the end of the point cloud processing pipeline to map the compact, regular grid back to its original point cloud form. Through theoretical and empirical analysis, we show that \textit{gridified networks} scale better in terms of memory and time than networks directly applied on raw point cloud data, while being able to achieve competitive results. Our code is publicly available at \url{https://github.com/computri/gridifier}.
\end{abstract}
\vspace{-9mm}
\section{Introduction}\label{sec:intro}
\vspace{-1mm}
\begin{figure}
  \centering
  \includegraphics[width=\linewidth]{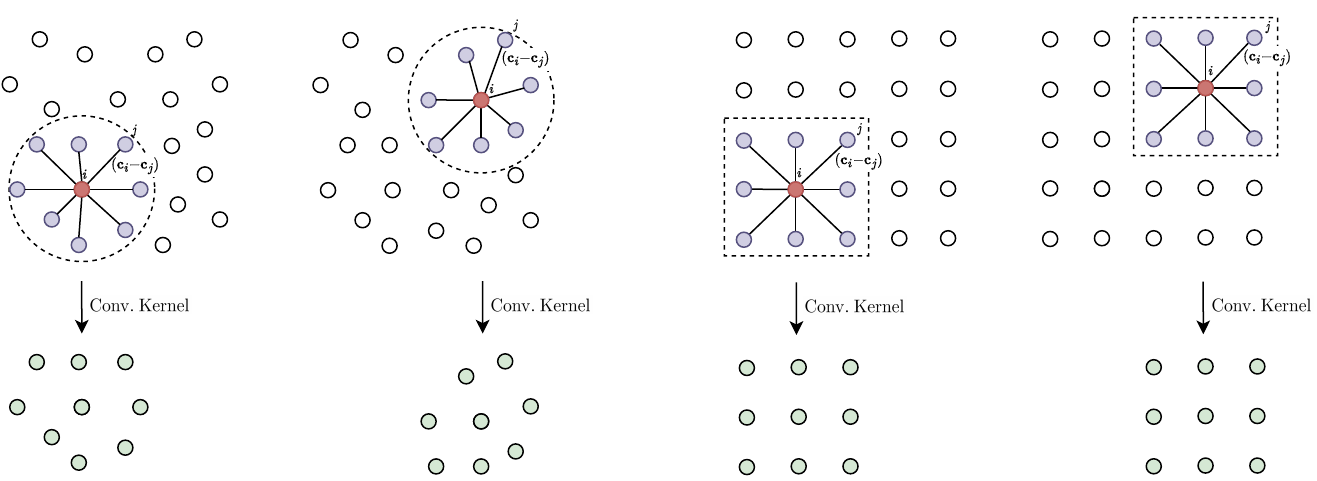}
  \vspace{-8mm}
  \caption{Convolution on point clouds and grids. Due to the irregular nature of point clouds, convolutional kernels --and other operations based on neighborhood information-- must be re-rendered for every query point in the point cloud (left). In contrast, grid data is regularly arranged, and thus pairwise distances are equal for any query point in the grid (right). As a result, the convolutional kernel can be computed once and reused for all query points.
  \vspace{-6mm}}
  \label{fig:convs}
\end{figure}
Point clouds provide sparse geometric representations of objects or surfaces equipped with signals defined over their structure, e.g., the surface normals of an underlying object \cite{wu20153d, qi2017pointnet} or the chemical properties of a molecule \cite{ramakrishnan2014quantum,schutt2017schnet}. Several neural operators have been developed that can be applied to such sparse representations provided by point clouds. These methods can be broadly understood as continuous generalizations of neural operators originally defined over regular discrete grids, e.g., convolution \cite{wu2019pointconv} and self-attention \cite{zhao2021point}.
\begin{figure*}
\centering
\hfill
\includegraphics[width=0.26\linewidth]{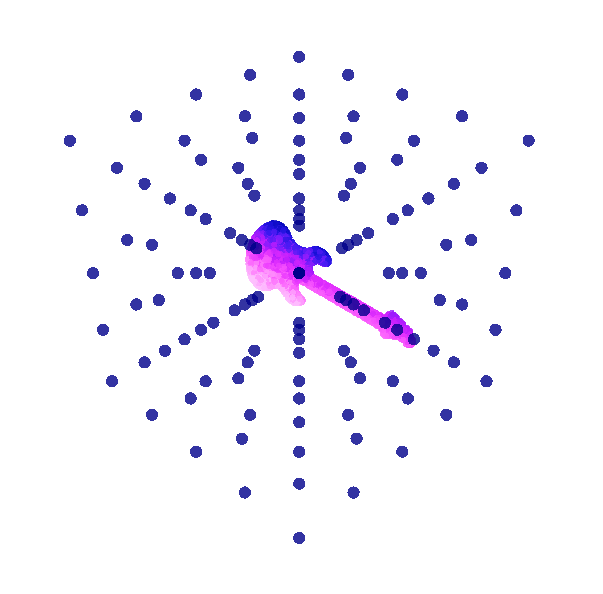}\hfill
\includegraphics[width=0.26\linewidth]{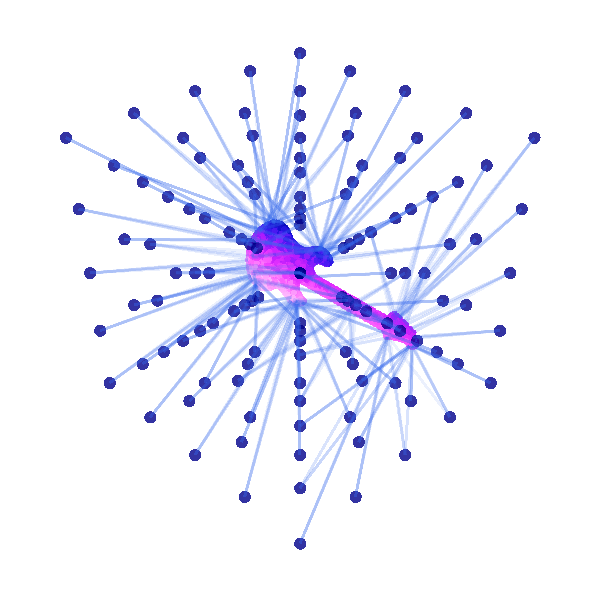}\hfill
\includegraphics[width=0.26\linewidth]{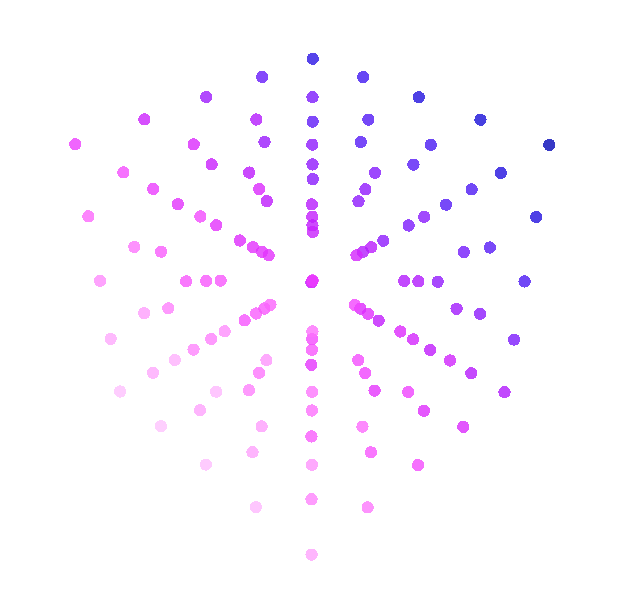}
\hfill
\vspace{-8mm}
\caption{Gridification. Gridification maps a point cloud $\gP$ onto a compact regular grid $\gG$. The method first constructs a $\Dt$-dimensional grid (left) that overlaps the point cloud. Then, it connects points on the point cloud to points in the grid given by a connectivity scheme $\gE_{\gP \rightarrow \gG}$, i.e., a set of edges from points in the point cloud to points in the grid,  determined by \textit{bilateral k-nearest neighbors connectivity} (middle). Finally, gridification propagates information from the point cloud onto the grid through a \textit{convolutional message passing} layer acting over the bipartite graph $(\gP, \gG,  \gE_{\gP \rightarrow \gG})$. By carefully selecting the different components of the gridification module, gridification is able to construct compact rich grid representations that can be subsequently processed with grid operations such as \texttt{Conv3D}.
\vspace{-3mm}}
\label{fig:gridification}
\end{figure*}

\textbf{The problem of learning on raw point clouds.} Unfortunately, the flexibility required from neural operators to accommodate irregular sparse representations like point clouds brings about important increases in time and memory consumption. This is especially prominent in neural operations that construct feature representations based on neighborhood information, e.g., convolution. In the case of point clouds, the irregular distances between points make these neural operations significantly more computationally demanding compared to regular grid representations like images or text. For instance, for convolution, the convolutional kernel needs to be recalculated for each point in a point cloud to account for irregular distances from the query point to other points in its neighborhood (Fig.~\ref{fig:convs} left). In contrast, grid representations standardize pairwise distances following a grid structure (Fig.~\ref{fig:convs} right). As a result, the distances from a point to all other points in its neighborhood are fixed for all points queried in the grid. Therefore, it is possible to compute the kernel once, and reuse it across all query positions. This difference illustrates that operations relying on neighborhood information scale much worse in terms of memory and time for point clouds than for grid data, specially for large inputs and large neighborhoods.
\begin{figure}
  \centering
  \includegraphics[width=\linewidth]{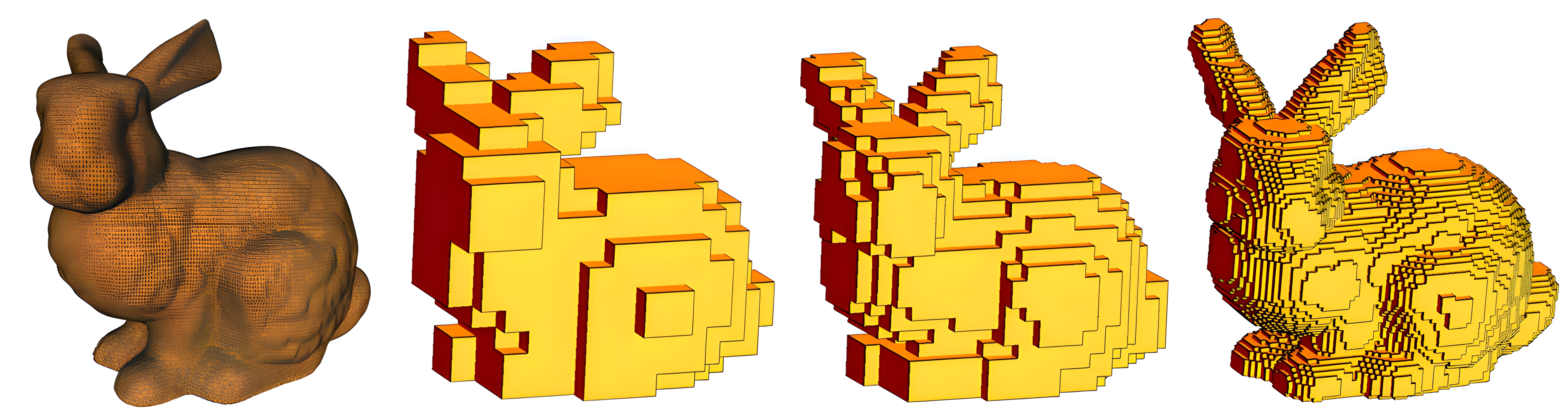}
  \vspace{-10mm}
  \caption{Voxelization of the Stanford Bunny \cite{turk1994zippered} for different resolutions. Taken from \citet{karmakar2011construction}.
  \vspace{-10mm}}
  \label{fig:voxelization}
\end{figure}

\textbf{A potential solution: Voxelization.} A potential solution to address the challenges posed by point clouds lies in treating the point cloud as a continuous density that can be sampled on a dense regular grid: a process called \textit{voxelization} \cite{maturana2015voxnet, wu20153d}. The idea of voxelization is to create a grid that overlaps with the domain of the point cloud (Fig.~\ref{fig:voxelization}). Although voxelization methods create grid representations on which neural operations defined on grids can act, e.g., \texttt{Conv3D}, the grids resulting from voxelization are oftentimes much larger than the number of points in the original point cloud. This is a consequence of (\textit{i}) the high-resolution grids required to describe fine details from the point cloud, and \textit{(ii)} the low occupancy of the grid resulting from the sparse nature of point clouds which generally leads to many more points to process in the resulting grid than in the original point cloud.

\textbf{Our proposed solution: Gridification.} In this paper, we propose an alternative solution to address the memory and computational scalability of point cloud methods by addressing its root cause: \textit{the irregularity of the data}. We propose \textit{learnable gridification} as the first step in a point cloud processing pipeline to transform the point cloud into a \textit{compact, regular grid} (Fig.~\ref{fig:gridification}). Thanks to gridification, subsequent layers can use operations defined on grids, e.g., \texttt{Conv3D}, which scale much better than native point cloud methods. In a nutshell, gridification can be understood as a \textit{convolutional message passing} layer acting on a \textit{bipartite graph} that establishes connections between points in the point cloud to points in the grid given by a \textit{bilateral $k$-nearest neighbor connectivity}. The proposed bilateral $k$-nearest neighbor connectivity guarantees that all points both in the point cloud and in the grid are connected, therefore allowing for the construction of expressive yet compact grid representations.

In contrast to voxelization, gridification produces expressive compact grid representations in which the number of points in the resulting compact regular grid is roughly equal to the number of points in the original point cloud, yet the grid is able to preserve fine geometric details from the original point cloud. For instance, we observe that point clouds with $\Nt{=}1000$ points can be effectively mapped to a compact dense $\mathrm{10x10x10}$ grid without significant information loss. We show through theoretical and empirical analysis that the resulting grid representations scale much better in terms of memory and time than native point cloud methods. This is verified on several comparison studies for increasing number of points in the point cloud and increasing neighborhood sizes in the construction of convolutional kernels.

We demonstrate that gridification can also be used for tasks from point clouds to point clouds, e.g., segmentation. To this end, we introduce a \textit{learnable de-gridification} step at the end of the point cloud processing pipeline, which can be seen as an inverted gridification step that maps the compact, regular grid back to its original point cloud form. This extension allows for the construction of \textit{gridified networks} --networks that operate on grids-- to solve global prediction tasks, e.g., classification, as well as dense prediction tasks, e.g., segmentation and regression, on point cloud data.


\vspace{-3mm}
\section{Method}\label{sec:method}
\vspace{-1mm}
\subsection{Point cloud and grid representations} 
\vspace{-1mm}
\textbf{Point cloud.} A point cloud $\gP {=}\{(\cv^\gP_i, \xv^\gP_i)\}_{i{=}1}^{\Nt_\gP}$ is an \textit{unstructured} set of $\Nt_\gP$ pairs of coordinate-feature values $(\cv^\gP_i, \xv^\gP_i)$ scattered in space without any predefined pattern or connectivity. Point clouds sparsely represent geometric structures through pairs of coordinate vectors $\cv^\gP_i \in \sR^\Dt$ and corresponding function values over that geometric structure $\xv_i^\gP \in \mathbb{R}^{\Ft_{\gP}}$, e.g., surface normals, RGB-values, electric potentials, etc.

\textbf{Grid.} A grid $\gG{=}\{(\cv^\gG_i, \xv^\gG_i)\}_{i{=}1}^{\Nt_\gG}$ can be interpreted as a point cloud on which the coordinate-feature pairs $(\cv^\gG_i, \xv^\gG_i)$ are arranged in a regular pattern that form a lattice. In contrast to general point clouds, points in a grid are evenly spaced and align along predefined axes, e.g., $x$, $y$, $z$. The regular spacing between points leads to regular pairwise distances for all query points in the grid. As a result, we can calculate pairwise attributes once, and reuse them for all query points. 

\vspace{-3mm}
\subsection{Gridification: From a point cloud to a dense grid}
\vspace{-1mm}
We seek to map the sparse point cloud $\gP {=}\{(\cv^\gP_i, \xv^\gP_i)\}_{j{=}1}^{\Nt_\gP}$ onto a compact regular grid $\gG {=}\{(\cv_i^\gG, \xv_i^\gG)\}_{i{=}1}^{\Nt_\gG}$ in $\sR^\Dt$. We formalize this process as an operation over a \textit{bipartite graph} that establishes connections between points in the point cloud $\gP$ to points in the grid $\gG$ given by a \textit{connectivity scheme} $\gE_{\gP \rightarrow \gG}$ defined as a set of edges $\ev_{j \rightarrow i} \in \gE_{\gP \rightarrow \gG}$.

\textbf{Learnable gridification as message passing.} We aim to learn a mapping from $\gP$ to $\gG$ such that the grid representation $\gG{=}\{(\cv^\gG_i, \xv^\gG_i)\}_{i{=}1}^{\Nt_\gG}$, $\cv_i \in \sR^\Dt$, $\xv^\gG_i \in \mathbb{R}^{\Ft_{\gG}}$, adequately represents the source point cloud $\gP$ for the downstream task. Given a source point cloud $\gP$, a target grid $\gG$ and a connectivity scheme $\gE_{\gP \rightarrow \gG}$, we define gridification as a \textit{convolutional message passing} layer \cite{gilmer2017neural} on the bipartite graph $(\gP, \gG, \gE_{\gP \rightarrow \gG})$ defined as:
\begin{equation}
    \xv^\gG_i = \phi_{\mathrm{upd}} \left( \bigoplus_{\ev_{j \rightarrow i} \ \in\  \gE_{\gP \rightarrow \gG}}\phi_{\mathrm{msg}}\Big(\phi_{\mathrm{node}}\left(\xv^\gP_j\right),  \phi_{\mathrm{pos}}\left(\cv^\gG_i - \cv^\gP_j\right) \Big) \right). 
\end{equation}
It consists of a node embedding network $\phi_\mathrm{node}: \sR^{\Ft_{\gP}} \rightarrow \sR^{\Ht}$ that processes the point cloud features $\xv_i^\gG$, a positional embedding network $\phi_\mathrm{pos}: \sR^{\Dt} \rightarrow \sR^{\Ht}$ that creates feature representations based on the pairwise distances between coordinates in $\gG$ and $\gP$ --thus resembling a convolutional kernel--, a message embedding network $\phi_\mathrm{msg}: \sR^{2\Ht} \rightarrow \sR^{\Ht}$ that receives both the node embedding and the relative position embedding to create the so-called \textit{message}. After the messages are created for all nodes described by connectivity of the node, these features are aggregated via the aggregation function $\bigoplus$, e.g., $\max$, $\mathrm{mean}$. Finally, the aggregated message is passed through the update network $\phi_\mathrm{upd}: \sR^{\Ht} \rightarrow \sR^{\Ft_{\gG}}$ to produce the grid feature representations $\xv_i^{\gG} \in \sR^{\Ft_{\gP}}$.
\vspace{-3mm}
\subsection{De-gridification: From a dense grid to a point cloud}
\vspace{-1mm}
To extend the use of gridification to tasks from the point cloud $\gP$ to the point cloud $\gP$, e.g., segmentation, regression, we define a \textit{de-gridification} step that sends a grid representation $\gG$ back to its original point cloud form $\gP$. Formally, the de-gridification step is defined as:
\begin{equation}
    \xv^\gP_{i} = \phi_{\mathrm{upd}} \left( \bigoplus_{\ev_{j \rightarrow i} \ \in\  \gE_{\gG \rightarrow \gP}}\phi_{\mathrm{msg}}\Big( \phi_{\mathrm{node}}\left(\xv^\gG_j\right),  \phi_{\mathrm{pos}}\left(\cv^\gP_i - \cv^\gG_j\right) \Big) \right). 
\end{equation}
Intuitively, de-gridification can be interpreted as a gridification step from $\gG$ to $\gP$ given by an inverted connectivity scheme $\gE_{\gG \rightarrow \gP} {=} ( \gE_{\gP \rightarrow \gG})^{-1}$. Note that, it is not necessary to calculate the connectivity scheme for the de-gridification step. Instead, we can obtain it simply by taking the connectivity scheme from the gridification step $\gE_{\gG \rightarrow \gP}$ and inverting the output and input nodes of the edges.
\vspace{-3mm}
\subsection{Requirements and properties of gridification}\label{sec:requirements}
\vspace{-1mm}
We desire to construct a compute and memory efficient grid representation $\gG$ that captures all aspects of the point cloud $\gP$ as good as possible. That is, a compact, yet rich grid representation $\gG$ that preserves the structure of the point cloud $\gP$ with as low loss of information as possible. With this goal in mind, we identify the following requirements:
\begin{enumerate}[label=(\textit{\roman*}), topsep=0pt, leftmargin=*, itemsep=0pt]
\item The number of points in the grid $\Nt_\gG$ should be at least as large as the number of points in the point cloud $\Nt_\gP$.\break
\vspace{-4mm}
\item The width of all hidden representations of the node embedding network $\phi_\mathrm{node}$ should be \textit{at least as large} as the width of the point cloud features $\xv_i^\gP$, i.e., $\Ft_{\gP}$. 
\item The width of all hidden representations of the position embedding network $\phi_\mathrm{pos}$ should be at least as large as the dimension of the domain $\Dt$.
\item The width of all hidden representations of the embedding networks $\phi_\mathrm{upd}$, $\phi_\mathrm{msg}$ should be \textit{at least as large} as the width of the point cloud features $\xv_i^\gP$ plus the dimension of the domain $\Dt$.
\item Each point $\cv^\gP$ in the point cloud should be connected to \textit{at least} one point $\cv^\gG$ in the grid. 
\item The positional embedding network $\phi_\mathrm{pos}$ should be able to describe high frequencies. 
\item Each point $\cv^\gG$ in the grid should be connected to \textit{at least} one point $\cv^\gP$ in the point cloud. 
\end{enumerate}
\textbf{Preventing information loss.} To prevent information loss, we want to avoid any kind of compression either in the grid representation or in any intermediary representation during the gridification process. Consequently, we restrict the number of points as well as the width of all representations to be at least as big as the corresponding dimensions in the source point cloud $\gP$ --items (\textit{i})-(\textit{iv})--. In addition, we must make sure that all points in the point cloud are connected to points in the grid to prevent points from being disregarded during gridification --item (\textit{v})--. Finally, we must also make sure that the positional embedding network $\phi_\mathrm{pos}$ is able to represent high frequencies --item (\textit{vi})--. This is important as multilayer perceptrons ($\mathrm{MLP}$s) with piecewise nonlinearities, e.g., $\mathrm{ReLU}$, have been shown to have an implicit bias towards smooth functions \cite{tancik2020fourier, sitzmann2020implicit}. In the context of gridification, this means that using conventional $\mathrm{MLP}$s for the positional embedding network $\phi_\mathrm{pos}$ could result in over-smooth grid representations unable to represent fine details from the source point cloud. We circumvent this issue by using parameterizations for $\phi_\mathrm{pos}$ able to model high frequencies (Sec.~\ref{sec:pos_network}).

\textbf{Encouraging compact representations.} In addition to encouraging no information loss, we also identify requirements that encourage the resulting grid representation to be compact and expressive. First, we note that item (\textit{v}) is important for this end as well, as over-smooth representations implicitly require higher resolutions to be able to encode fine-grained details. Additionally, we impose all points in the grid to be connected to points in the point cloud --item (\textit{vii})-- to prevent the grid representation from having low occupancy. This restriction allows us to make sure that all the spatial capacity of the grid is being used. This in turn allows us to construct compact rich grid representations.
\vspace{-3mm}
\subsection{Materializing the gridification module}
\vspace{-1mm}
Based on the previous requirements and properties, we define the components of the gridification module as follows:

\vspace{-2mm}
\subsubsection{The grid $\gG$}
\vspace{-1mm}
 Let $[a, b]^\Dt$ be the domain of the point cloud $\gP$, i.e., $\cv^\gP_i \in [a, b]^{\Dt}$, $\forall\ \cv^\gP_i\in \gP$. Then, we define the regular grid $\gG$ over the same domain $[a, b]^\Dt$ with $\sqrt[\leftroot{-3}\uproot{3}\Dt]{\Nt^{\gG}}$ points along each dimension. By doing so, we guarantee that the grid $\gG$ is uniformly spaced over the domain of the point cloud, therefore (\textit{i}) preserving the statistics of the input point cloud, and (\textit{ii}) being able to represent the underlying signal in the same range. In practice, point clouds are normalized during the preprocessing steps preceding a point cloud processing pipeline. As a result, we often have that $a{=}-1$ and $b{=}1$, leading to a point cloud and a grid defined on $[-1, 1]^\Dt$.
 \vspace{-2mm}
\subsubsection{The connectivity scheme $\gE_{\gP \rightarrow \gG}$}
\begin{figure}
    \centering
    \includegraphics[width=0.75\textwidth]{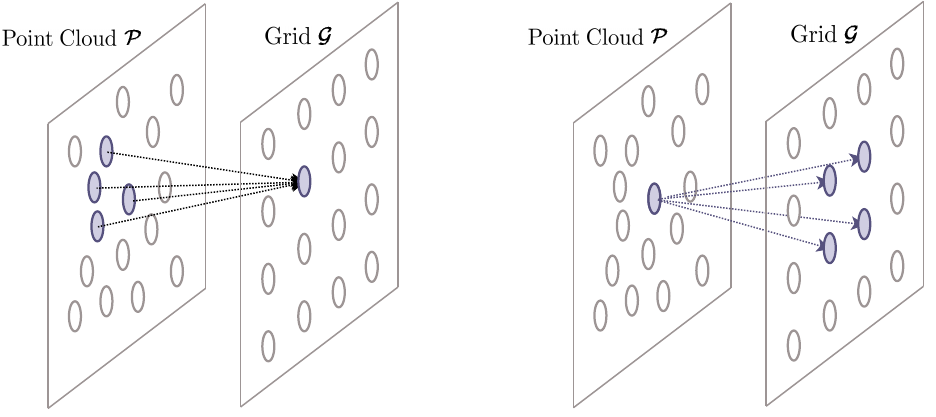}
    \vspace{-6mm}
    \caption{Bilateral $k$-nearest neighbor connectivity for $k{=}4$.
    \vspace{-4mm}}
    \label{fig:bilateral_conectivity}
\end{figure}

\vspace{-1mm}
Motivated by the requirements in Sec.~\ref{sec:requirements}, we opt for \textit{bilateral $k$-nearest neighbor connectivity} over common alternatives such as radius connectivity \cite{qi2017pointnet, qi2017pointnet++} or one-way $k$-nearest neighbor connectivity \cite{barber1996quickhull, connor2010fast} for the construction of the connectivity scheme $\gE_{\gP \rightarrow \gG}$ to guarantee that no points either in the grid $\gG$ nor the point cloud $\gP$ are disconnected. Bilateral $k$-nearest neighbor connectivity consists of a two-way $k$-nearest neighbor approach in which first each point $\cv_i^{\gG}$ in the grid is linked to the $k$ nearest points $\cv_j^\gP$ in the point-cloud. Subsequently, connections are established from each point $\cv_i^{\gP}$ in the point cloud to its nearest $k$ points $\cv_j^{\gG}$ in the grid (Fig.~\ref{fig:bilateral_conectivity}). By following this procedure, bilateral $k$-nearest neighbor connectivity creates a \textit{complete} connectivity scheme, i.e., with no disconnected points, from $\gP$ to $\gG$ with at least $k$ and at most $2k$ connections for each point.
\vspace{-2mm}
\subsubsection{The positional embedding network $\phi_\mathrm{pos}$}\label{sec:pos_network}
\vspace{-1mm}
In literature, the positional embedding network $\phi_\mathrm{pos}$ is often parameterized as an $\mathrm{MLP}$ with piecewise nonlinearities, e.g., $\mathrm{ReLU}$, that receives relative positions $(\cv_i {-} \cv_j)$ as input and retrieves the value of an spatial function at that position $\phi_\mathrm{pos}(\cv_i {-} \cv_j)$ \cite{schutt2017schnet, qi2017pointnet++, wu2019pointconv}. However, previous studies have shown that $\mathrm{MLP}$s with piecewise nonlinearities suffer from an spectral bias towards low frequencies, which limits their ability to represent functions with high frequencies \cite{tancik2020fourier, sitzmann2020implicit}. In the context of modelling spatial neural operators such as $\phi_\mathrm{pos}$, this implies that using piecewise $\mathrm{MLP}$s to parameterize spatial neural operators leads to inherently smooth operators. Consequently, applying such an operator over an input function, e.g., via a convolution operation, would implicitly perform a low-pass filtering of the input, causing the output representations to lack information regarding fine-grained details of the input.

To overcome this issue, we rely on the insights from \textit{Continuous Kernel Convolutions} \cite{romero2021ckconv} and parameterize the positional embedding network as a \textit{Neural Field} \cite{sitzmann2020implicit, tancik2020fourier}. In contrast to piecewise $\mathrm{MLP}$s, neural fields easily model high frequencies, and thus allow for powerful parameterizations of spatial neural operators that do not perform smoothing. In the context of gridification, using neural fields to parameterize $\phi_\mathrm{pos}$ allows gridification to project fine-grained geometric information from the point cloud onto the grid.
\vspace{-3mm}
\subsection{Gridified networks for global and dense prediction}
\label{section:gridnetworks}
\begin{figure}
    \centering
    \includegraphics[width=0.75\linewidth]{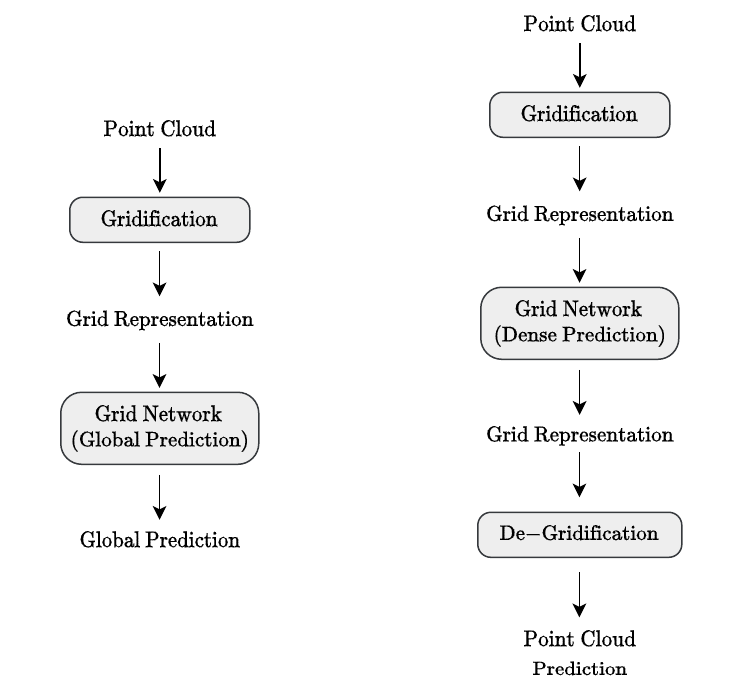}
    \vspace{-6mm}
    \caption{Point cloud processing pipeline for global prediction (left) and dense prediction tasks (right).
    \vspace{-4mm}}
    \label{fig:gridified_nets}
\end{figure}

\vspace{-1mm}
Gridification and de-gridification allow for the construction of \textit{gridified networks} able to process point clouds both for global and dense prediction tasks (Fig.~\ref{fig:gridified_nets}). For global prediction tasks, e.g., classification, we construct a point cloud processing pipeline consisting of gridification, followed by a \textit{grid network}, i.e., a neural network that operates on grid data, designed for global prediction, e.g., a ResNet \cite{he2016deep} or a ViT \cite{dosovitskiy2020image}. For dense prediction tasks, e.g., segmentation, our proposed point cloud pipeline consists of gridification, followed by a grid network designed for dense predictions, e.g., a U-Net \cite{ronneberger2015u} or a CCNN \cite{knigge2023modelling}. After the processed grid representation is obtained, we utilize the de-gridification step to map back the grid representation to a point cloud with the output node predictions.

\vspace{-3mm}
\section{Related Work}
\vspace{-1mm}
Deep learning approaches for point cloud processing can be broadly classified in two main categories: (\textit{i}) native point cloud methods and (\textit{ii}) voxelization methods.

\textbf{Native point cloud methods.} Native point cloud methods operate directly on the raw, irregular point cloud data without any preprocessing steps such as voxelization. These methods leverage the inherent spatial distribution of the points to extract meaningful features. PointNet \cite{qi2017pointnet} introduced a pioneering framework for point cloud processing by employing shared multilayer perceptrons and symmetric functions to learn global and local features from unordered point sets. PointNet++ \cite{qi2017pointnet++} extended this work with hierarchical neural networks to capture hierarchical structures in point clouds. PointConv \cite{wu2019pointconv} introduced a convolution operation specifically designed for point clouds, incorporating local coordinate systems to capture local geometric structures. PointGNN \cite{shi2020point} utilized graph neural networks to model interactions between neighboring points in point clouds. 

Despite the flexibility in handling irregular data that native point cloud methods provide, they suffer from scalability issues due to the increased computational and memory complexity of processing unstructured point sets.

\textbf{Voxelization methods.} Voxelization methods aim to convert the irregular point cloud data into a regular grid structure, enabling the utilization of neural architectures designed for regular grid data. VoxNet \cite{maturana2015voxnet} introduced the concept of voxelization for point clouds and employed 3D convolutions on the resulting grid representations. Volumetric CNN \cite{qi2016volumetric} extended this approach with an occupancy grid representation and achieved impressive performance on 3D shape classification tasks. Other works, such as VoxSegNet \cite{wang2019voxsegnet} explore variations of voxelization techniques to improve performance on tasks like object detection and segmentation. 

While voxelization methods offer a well-founded solution to the computational and memory complexity of native point cloud methods, in practice, they suffer from high memory consumption and information loss due to the discretization process. This is due to the inherent trade-off between the need to capture fine geometric details --which requires high resolution grids--, and the need for efficiency --which favors low resolution grids--. As a result, conventional voxelization methods struggle to strike a balance between resolution and speed. In contrast, gridification is able to generate compact yet expressive grid representations able to preserve fine geometric details on a low resolution grid with roughly the same number of points as the source point cloud. 

\textbf{Hybrid methods.} Aside from pure  point cloud and voxelization methods, there exist works that attempt combine the advantages of both categories. Their main idea is to combine point-wise and grid-wise operations to perform effective feature extraction while maintaining scalability and efficiency.\break PointGrid \cite{le2018pointgrid} uses a hybrid representation by voxelizing the point cloud and employing a combination of point-wise and grid-wise operations at each layer. Point-Voxel CNN \citep{liu2019point} combines grid convolutions with point-wise feature extraction. It uses low-resolution voxelization to aggregate neighborhoods with regular 3D convolutions and $\mathrm{MLP}$s to generate point-wise features that preserve fine-grained structure. These features are  then fused through interpolation. Point-Voxel Transformer\break \cite{zhang2022pvt} follows a similar two-branch structure, but replaces convolutions with windowed self-attention.

Although hybrid methods reduce the computational and memory complexity of native point cloud methods, their explicit use of voxelization still leads to a trade-off between information loss and efficiency on that branch. To compensate for the information lost during voxelization, they require a parallel raw point cloud branch, which does not scale well. In contrast, gridification does not make use of raw point cloud branches but instead focuses on the creation of descriptive compact grid representations that preserve the geometric information of the source point cloud. Hence, gridification offers a solution with better scalability properties than existing hybrid methods.  

\vspace{-3mm}
\section{Experiments}
\vspace{-1mm}
To evaluate our approach, we first analyze the expressive capacity of gridification and de-gridification on a toy point cloud reconstruction task. Next, we construct gridified networks and evaluate them on classification and segmentation tasks. In addition, we provide empirical analyses on the computational and memory complexity of gridified networks which we then corroborate with theoretical analyses. 
\begin{table*}[t]
\centering
    \begin{minipage}{\textwidth}
    \centering
    \caption{Classification performance on ModelNet40 benchmark.}
    \label{tab:modelnet_results}
    \vspace{-2.5mm}
    \begin{small}
    \scalebox{0.85}{
    \begin{tabular}{lllll}
    \toprule
    \sc{Model} & \sc{Input}  & \sc{Type} &  \sc{Accuracy} & \sc{Parameters} \\
    \midrule
    PointNet++ \cite{qi2017pointnet++} & $32 \times 1000$ & native & 89.64 &  1.5M \\
    VoxNet \cite{maturana2015voxnet} & $32\times 30^3$ & voxelization & 83.00 & 0.92M\\
    PointGrid \cite{le2018pointgrid} & $32\times16^3$ & voxelization & 92.00	 & - \\
    Point Voxel Transformer \cite{zhang2022pvt} & $32\times1024$ & hybrid & 94.00 & 2.76M\\
    Gridified Networks 3x3x3 (Ours) & $32\times1000 \rightarrow 32 \times 3^3$& voxelization & 90.86   & 0.28M  \\
    Gridified Networks 9x9x9 (Ours)  & $32\times1000 \rightarrow 32 \times 9^3 $ & voxelization & 92.28   & 0.47M  \\
    \bottomrule
    \end{tabular}}
    \end{small}
    \end{minipage}%
    \vspace{-3mm}
\end{table*}

\textbf{Experimental setup.} For the position embedding function $\phi_\mathrm{pos}$ we use an Random Fourier Feature Network \cite{tancik2020fourier}, due to explicit control over the smoothness through the initial frequency parameter $\Omega$. The practical setup and instantiation of the convolution blocks can be found in Appendix \ref{app:networkarch}. We train our models without data augmentation using AdamW \cite{loshchilov2018decoupled} and a cosine scheduler \citep{loshchilov2017sgdr} with 10 epochs of linear warm-up. We follow the standard procedure and preprocess all objects in the datasets to be centered and normalized. For each dataset, we choose the grid resolution such that its number of points is roughly equal to the size of the original point cloud. For ModelNet40 we use surface normals in addition to positions as node features. Dataset specific hyperparameters can be found in Appendix \ref{app:hyperparams}. 


\begin{figure}
    \centering
    \includegraphics[width=0.85\textwidth]{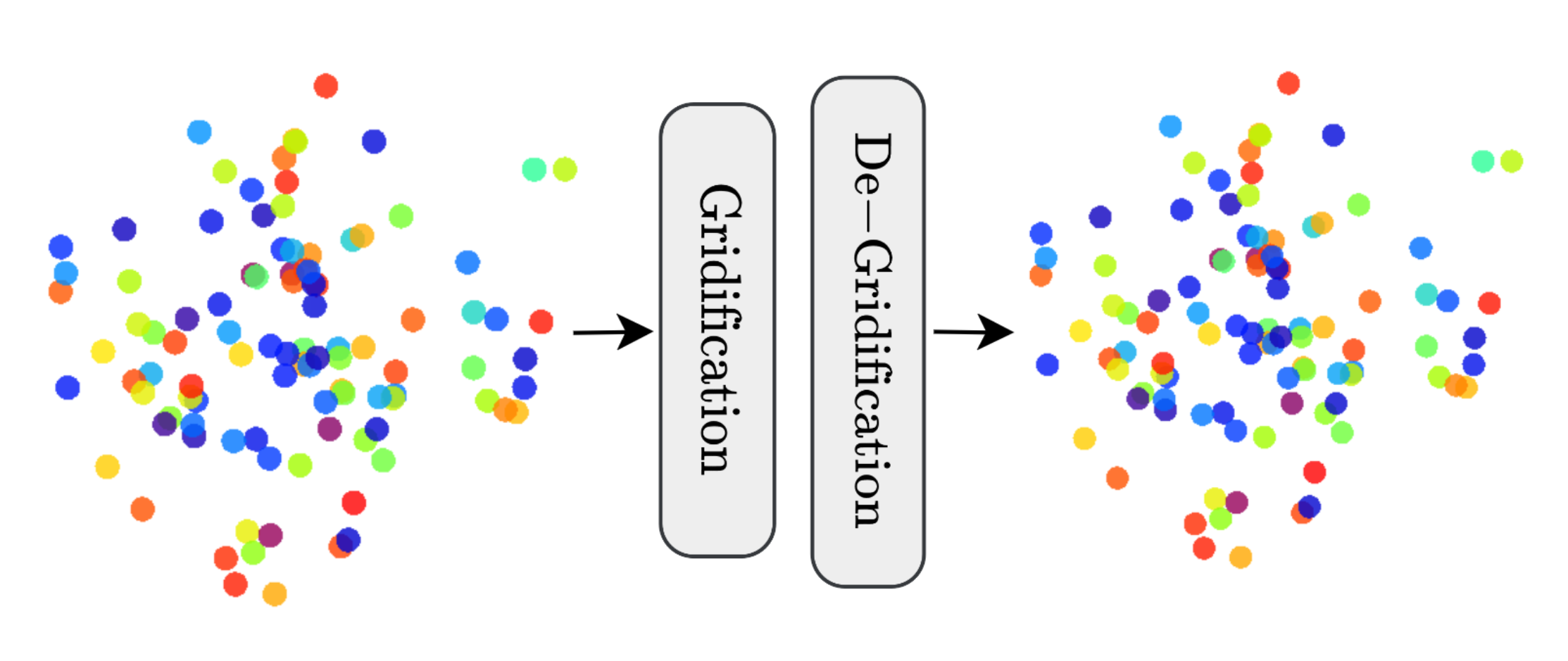}
    \vspace{-6mm}
    \caption{Random point clouds with random scalar node features are mapped to a grid representation. From the grid representation the node features need to be reconstructed via de-gridification.
    \vspace{-4mm}}
    \label{fig:recon_pipeline}
\end{figure}

\vspace{-3mm}
\subsection{Random point cloud reconstruction}
\vspace{-1mm}
First, we evaluate the expressivity of our proposed gridification and de-gridification procedure. To this end, we construct a dataset with 1000 synthetic random graphs --800 for training and 200 for validation-- consisting of a predefined number of nodes $\Nt^\gP{=}1000$ randomly sampled on the unit cube, i.e., $\cv_i^\gP \sim \gU( [-1, 1]^3)$, accompanied with a random scalar feature $f_i^\gP \sim \gU(-1, 1)$ at each position. 

\textbf{Experimental setup.} To evaluate the expressiveness of our method, we set up a network consisting only of a gridification and a de-gridification step, i.e., no intermediary layers, in a point cloud reconstruction pipeline. In other words, the task consists of propagating the point cloud into a grid representation, and mapping the grid representation back to the original point cloud (see Fig. \ref{fig:recon_pipeline}). 
Therefore, to successfully reconstruct the original point cloud from the grid representation, the grid representation must be able to retain sufficient information from the input point cloud. 

\begin{figure}
    \centering
    \includegraphics[width=0.85\textwidth]{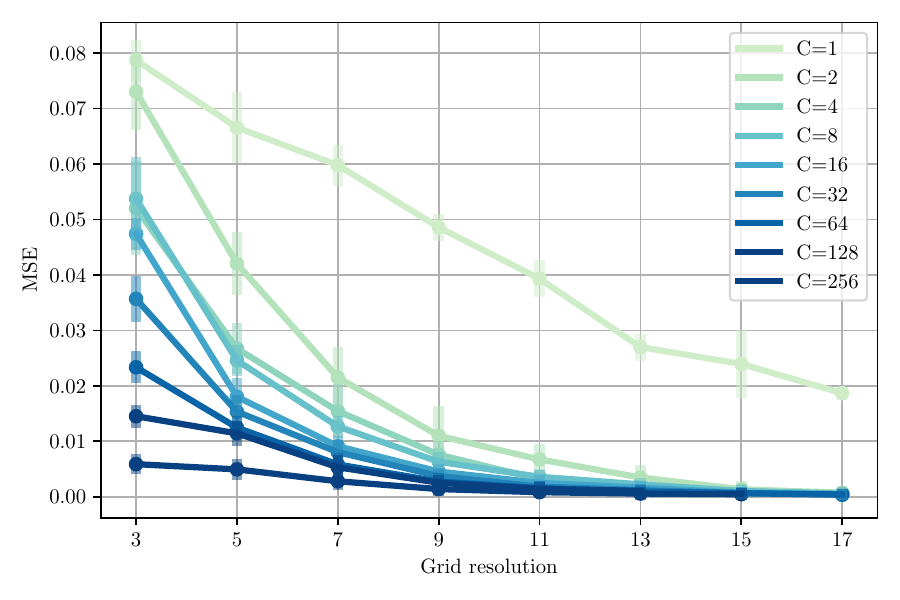}
    \vspace{-6mm}
    \caption{Random point cloud reconstruction error for varying grid resolution and number of channels on the grid representation.
    \vspace{-6mm}}
    \label{fig:recon}
\end{figure}

\textbf{Results.} Fig. \ref{fig:recon} shows reconstruction errors for different resolutions and different number of channels in the intermediary grid representation. We observe that it is possible to obtain good reconstructions by increasing the resolution of the grid or its number of channels. From an efficiency perspective, it is preferred to utilize low resolution representations with a larger number of channels due to the exponential growth in computational demands associated with higher grid resolutions, which instead scale linearly with the number of channels of the representation. Our experiments show that gridification is able to obtain compact grid representations that preserve the structure of the input point cloud. Furthermore, the quality of the grid representations can be efficiently improved by scaling the number of channels used. 
\vspace{-7mm}
\subsection{ModelNet40 classification} \label{sec:modelnet40exp}
\vspace{-1mm}
Next, we evaluate gridification on point cloud classification. We deploy gridified networks on ModelNet40 \cite{wu20153d}: a synthetic dataset for 3D shape classification, consisting of 12,311 3$\Dt$ meshes of objects belonging to 40 classes. ModelNet40 is broadly used as a point cloud benchmark in which points are uniformly sampled from the faces of the meshes.


\textbf{Results.} Our results (Tab.~\ref{tab:modelnet_results}) show that gridified networks achieve competitive performance while being significantly more efficient in terms of parameters, compute and memory. Interestingly, and in contrast to voxelization methods, we observe that gridified networks operate well even on extremely low resolution grids. For instance, on a $3{\times}3{\times}3$ grid, gridified networks attain an accuracy of $90.86\%$.
\vspace{-3mm}
\subsection{ShapeNet part segmentation} 
\vspace{-1mm}
Next, we evaluate gridification on point cloud segmentation. To this end, we deploy gridified networks on ShapeNet \cite{shapenet2016}: a synthetic dataset with 16,000 point clouds of objects from 16 categories, each of which contains 2 to 6 parts. The objective of the task is to segment the point clouds into one of 50 possible part annotations.



\textbf{Results.} Our results (Tab.~\ref{tab:shapenet_results}) demonstrate that gridified networks are also able to achieve competitive performance in segmentation tasks, while being significantly more efficient in terms of parameters, compute and memory. This result validates the ability of gridification to handle dense prediction tasks via gridification and de-gridification.
\begin{table}
\centering
    \begin{minipage}{\textwidth}
    \centering
    \caption{Segmentation performance on ShapeNet-part benchmark.}
    \label{tab:shapenet_results}
    \vspace{-2.5mm}
    \begin{small}
    \scalebox{0.85}{
    \begin{tabular}{l|lll}
    \toprule
    \sc{Model} & Gridified Networks  & PointNet++  & PointGrid \\
    \midrule
   \sc{Type} & voxelization & native & hybrid
    \\
     \midrule
     instance average IoU & 87.07 & 85.1 & 86.4\\
     class average IoU & 81.68 & 81.9 & 82.2\\
     \midrule
airplane & 88.52  & 82.4 & 85.7\\
bag & 86.54  &  79.0 & 82.5 \\
cap & 74.09  & 87.7 & 81.8 \\
car & 80.46  & 77.3 & 77.9 \\
chair & 91.44  & 90.8 & 92.1 \\
earphone & 51.81 & 71.8 &  82.4 \\
guitar & 92.61 &  91.0 &  92.7 \\
knife & 89.44  & 85.9 & 85.8 \\
lamp & 82.07  &  83.7 & 84.2 \\
laptop & 96.07  &  95.3 & 95.3 \\
motor & 65.36  & 71.6 & 65.2 \\
mug & 92.99  &  94.1 & 93.4 \\
pistol & 86.72 & 81.3  & 81.7 \\
rocket & 58.57   & 58.7 & 56.9 \\
skateboard & 75.70   & 76.4 & 73.5 \\
table & 85.66 &  82.6 & 84.6 \\
    \bottomrule
    \end{tabular}}
    \end{small}
    \end{minipage}%
    \vspace{-4mm}
\end{table}
\vspace{-3mm}
\subsection{Efficiency analysis of gridification} 
\vspace{-1mm}
Finally, we investigate the scalability properties of gridification. Specifically, we analyze the time and memory consumption of gridified networks during inference on ModelNet40 for point clouds with increasing size, and compare the computation and memory complexity of convolutional operations on grid and point cloud data.

\begin{figure*}
\centering
\hfill
\includegraphics[width=0.45\linewidth]{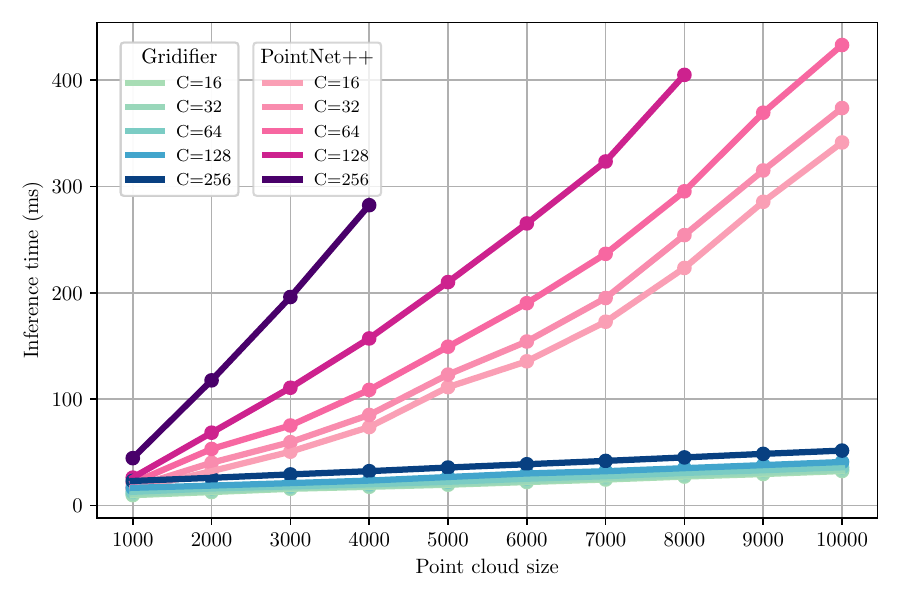}\hfill
\qquad
\includegraphics[width=0.45\linewidth]{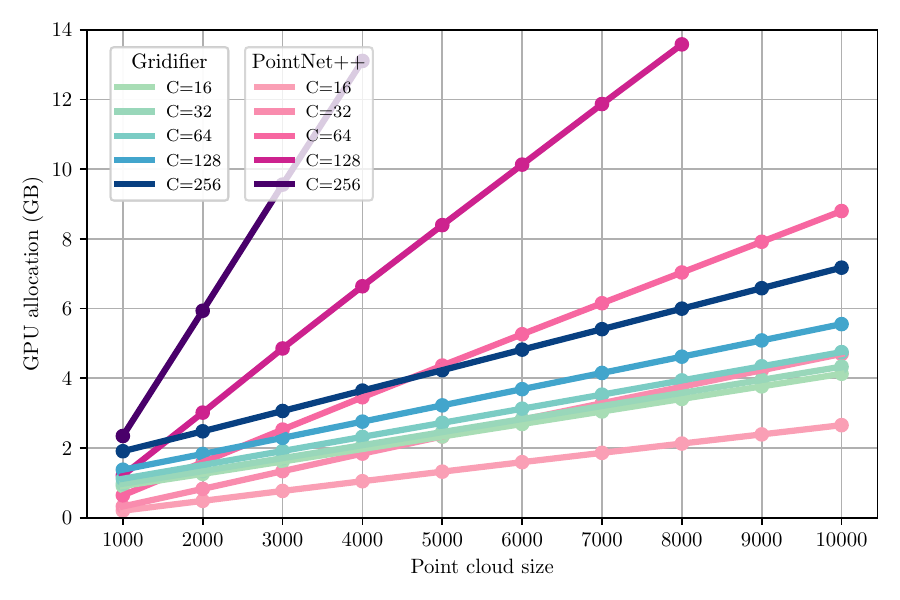}\hfill
\caption{Average time (left) and GPU allocation (right) during inference on ModelNet40 for a batch size of $32$.}
\vspace{-8mm}
\label{fig:efficiency_modelnet}
\end{figure*}

\textbf{Scaling gridified networks to large point clouds.} Fig~\ref{fig:efficiency_modelnet} shows the average time and memory consumption during inference on ModelNet40 for gridified networks and PointNet++. We observe that gridified networks exhibit a much more favorable scalability both in terms of inference time and GPU allocation --linear vs. quadratic-- as the input size and number of channels increase. This demonstrates that gridified networks scale much better than native point cloud methods both for larger point clouds and larger networks. 

\textbf{Scaling the receptive field of neural operations.} Furthermore, we analyze the scalability properties of gridified networks relative to the size of its receptive fields. As illustrated in Fig.~\ref{fig:convs}), for native point cloud methods the convolutional kernel must be recomputed for all query points in the point cloud. As a consequence, the construction of the convolutional kernels of size $\Kt$ for all query points in a point cloud with $\Nt$ points incurs in $\gO(\Kt \Dt)$ memory and time complexity. In contrast, on grid data, we can compute the kernel once and reuse it at all positions. As a result, on a grid, this operation incurs in $\gO(\Dt)$ time and memory complexity.

\begin{figure*}
\centering
\hfill
\includegraphics[width=0.44\linewidth]{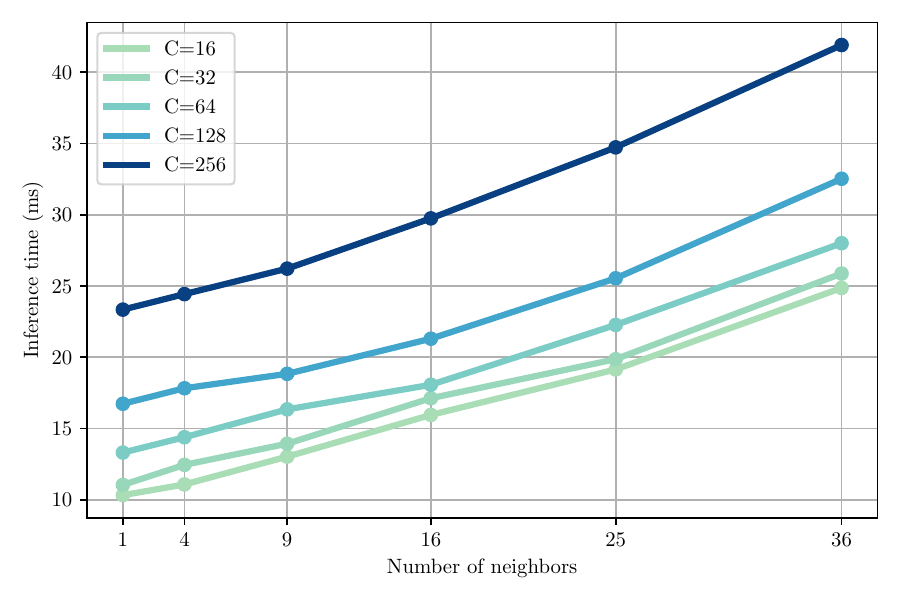}\hfill
\qquad
\includegraphics[width=0.47\linewidth]{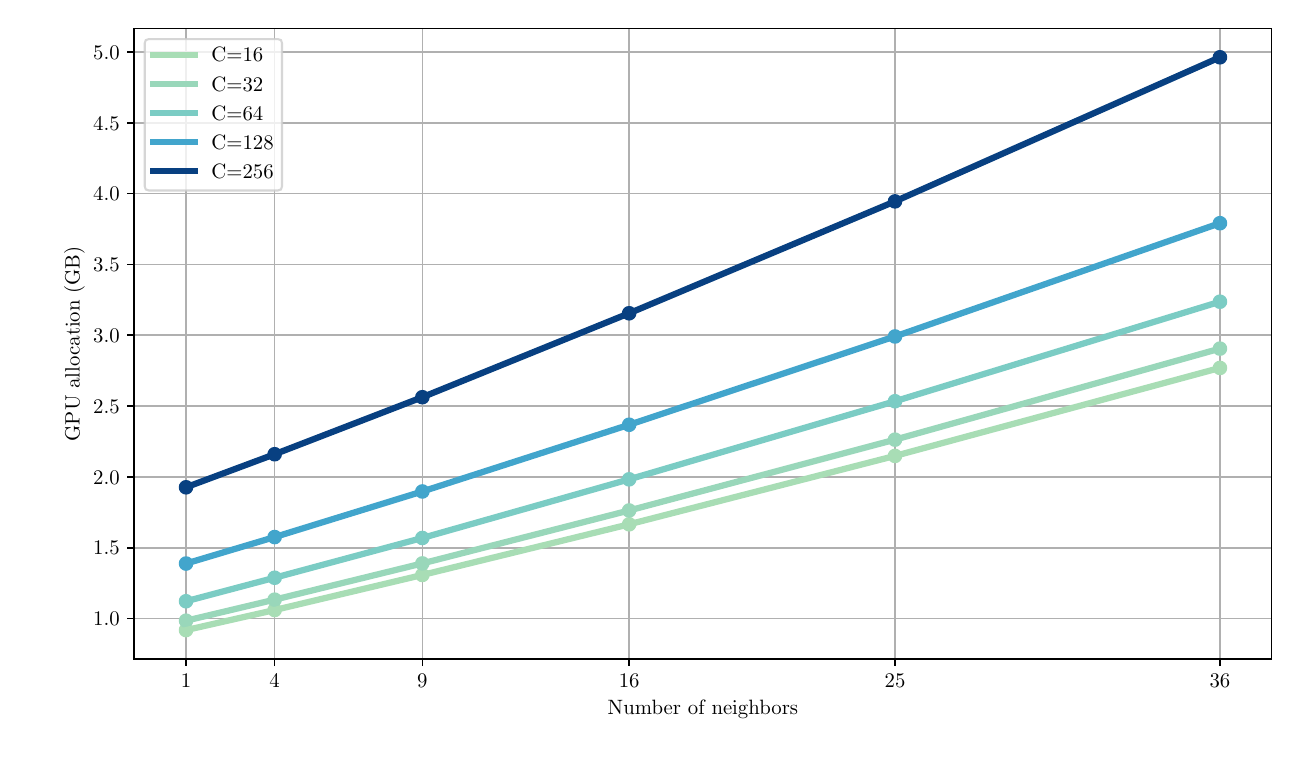}\hfill
\caption{Average time (left) and GPU allocation (right) on ModelNet40 validation set per batch $B=32$ for various number of neighbors and number of channels $C$ on the grid representation.}
\vspace{-8mm}
\label{fig:efficiency_neighbors}
\end{figure*}

Fig. \ref{fig:efficiency_neighbors} show the methods' potential to scale up the receptive field of the gridification module without introducing significant computational overhead.

\vspace{-3mm}
\section{Limitations and future work}
\vspace{-1mm}
\textbf{The resolution of gridification depends on the size of the point cloud.} The main limitation of gridification is that the resolution on the grid is directly proportional to the size of point cloud in order to preserve information. This in turn means that the whole gridified architecture must be changed for point clouds of different sizes, even if they represent the same underlying signal. This is in contrast to native point cloud methods, which, due to their continuous nature, are, in principle, able to generalize to point clouds of different sizes as long as these exhibit the same structures.

\textbf{Towards no information loss.}  While gridification aims to produce compact grid representations with minimal information loss, our experiments reveal that some information still gets lost in the process. Loosely speaking, it should be possible to create grid representations that do not lose any information by ensuring that the grid representation has at least as many points and as many channels as the source point cloud representation. Gaining richer theoretical understanding of gridification, could therefore lead to grid representations with no information loss either by imposing other requirements on gridification, or by considering different functional families in the gridification process.

\textbf{Large scale point clouds and global context.} While we verify the scalability and efficiency of gridified networks for increasing point cloud sizes, we only carry on experiments on relatively small datasets. In future work, we aim to deploy gridification to large scale datasets. Furthermore, recent works have shown that using global receptive fields in convolutional operations consistently leads to better results across several tasks, even outperforming well-established Transformer architectures \cite{gu2021efficiently, knigge2023modelling, poli2023hyena}. Due to the computational complexity of native point cloud methods, networks with global context have not been explored for point cloud processing. With gridification this ability becomes computationally feasible. Exploring the effect of global context for point cloud processing is an exciting research direction.

\textbf{Generative tasks.} Gridification opens up the possibility of performing scalable generative tasks on large point clouds. Gridification can directly be extended to generative tasks if we assume that the point-cloud structure is preserved, i.e., if the coordinates of the output and input point clouds are equal. If this is not the case, e.g., for the generation of molecules \cite{xu2019deep, hoogeboom2022equivariant}, de-gridification module must be modified to predict both the features and positions of the new point cloud. We consider this a particularly promising direction for future research. 

\textbf{Equivariant gridification.} In its current form, gridified networks do not respect symmetries which might be important for some applications, e.g., equivariance to 3D rotations for the prediction and generation of molecules \cite{schutt2017schnet, hoogeboom2022equivariant}. In future work, we aim to extend gridification to respect these symmetries by taking inspiration from equivariant graph neural networks \cite{fuchs2020se, satorras2021n}. It is important to note that not only gridification and de-gridification must be equivariant, but also that the grid operations in between should respect these properties. This can be achieved in an efficient yet expressive manner through the use of continuous Monte-Carlo convolutions on the regular representations of the group \cite{finzi2020generalizing, romero2022learning}.


\vspace{-3mm}
\section{Conclusion}
\vspace{-1mm}
This work presents gridification, a method that strongly reduces the computational requirements of point cloud processing pipelines by mapping input point clouds to a grid representation, and performing neural operations in there. We demonstrate that gridified networks are able to match the accuracy of native point cloud methods, while being much faster and memory efficient. Through empirical and theoretical analyses, we also show that gridified networks scale much more favorably than native point cloud methods to larger point clouds and larger neighborhoods.
\newpage



\bibliography{references}

\begin{thebibliography}{37}
\providecommand{\natexlab}[1]{#1}
\providecommand{\url}[1]{\texttt{#1}}
\expandafter\ifx\csname urlstyle\endcsname\relax
  \providecommand{\doi}[1]{doi: #1}\else
  \providecommand{\doi}{doi: \begingroup \urlstyle{rm}\Url}\fi

\bibitem[Barber et~al.(1996)Barber, Dobkin, and Huhdanpaa]{barber1996quickhull}
Barber, C.~B., Dobkin, D.~P., and Huhdanpaa, H.
\newblock The quickhull algorithm for convex hulls.
\newblock \emph{ACM Transactions on Mathematical Software (TOMS)}, 22\penalty0
  (4):\penalty0 469--483, 1996.

\bibitem[Connor \& Kumar(2010)Connor and Kumar]{connor2010fast}
Connor, M. and Kumar, P.
\newblock Fast construction of k-nearest neighbor graphs for point clouds.
\newblock \emph{IEEE transactions on visualization and computer graphics},
  16\penalty0 (4):\penalty0 599--608, 2010.

\bibitem[Dosovitskiy et~al.(2020)Dosovitskiy, Beyer, Kolesnikov, Weissenborn,
  Zhai, Unterthiner, Dehghani, Minderer, Heigold, Gelly,
  et~al.]{dosovitskiy2020image}
Dosovitskiy, A., Beyer, L., Kolesnikov, A., Weissenborn, D., Zhai, X.,
  Unterthiner, T., Dehghani, M., Minderer, M., Heigold, G., Gelly, S., et~al.
\newblock An image is worth 16x16 words: Transformers for image recognition at
  scale.
\newblock \emph{arXiv preprint arXiv:2010.11929}, 2020.

\bibitem[Finzi et~al.(2020)Finzi, Stanton, Izmailov, and
  Wilson]{finzi2020generalizing}
Finzi, M., Stanton, S., Izmailov, P., and Wilson, A.~G.
\newblock Generalizing convolutional neural networks for equivariance to lie
  groups on arbitrary continuous data.
\newblock In \emph{International Conference on Machine Learning}, pp.\
  3165--3176. PMLR, 2020.

\bibitem[Fuchs et~al.(2020)Fuchs, Worrall, Fischer, and Welling]{fuchs2020se}
Fuchs, F., Worrall, D., Fischer, V., and Welling, M.
\newblock Se (3)-transformers: 3d roto-translation equivariant attention
  networks.
\newblock \emph{Advances in Neural Information Processing Systems},
  33:\penalty0 1970--1981, 2020.

\bibitem[Gilmer et~al.(2017)Gilmer, Schoenholz, Riley, Vinyals, and
  Dahl]{gilmer2017neural}
Gilmer, J., Schoenholz, S.~S., Riley, P.~F., Vinyals, O., and Dahl, G.~E.
\newblock Neural message passing for quantum chemistry.
\newblock In \emph{International conference on machine learning}, pp.\
  1263--1272. PMLR, 2017.

\bibitem[Gu et~al.(2021)Gu, Goel, and R{\'e}]{gu2021efficiently}
Gu, A., Goel, K., and R{\'e}, C.
\newblock Efficiently modeling long sequences with structured state spaces.
\newblock \emph{arXiv preprint arXiv:2111.00396}, 2021.

\bibitem[He et~al.(2016)He, Zhang, Ren, and Sun]{he2016deep}
He, K., Zhang, X., Ren, S., and Sun, J.
\newblock Deep residual learning for image recognition.
\newblock In \emph{Proceedings of the IEEE conference on computer vision and
  pattern recognition}, pp.\  770--778, 2016.

\bibitem[Hoogeboom et~al.(2022)Hoogeboom, Satorras, Vignac, and
  Welling]{hoogeboom2022equivariant}
Hoogeboom, E., Satorras, V.~G., Vignac, C., and Welling, M.
\newblock Equivariant diffusion for molecule generation in 3d.
\newblock In \emph{International Conference on Machine Learning}, pp.\
  8867--8887. PMLR, 2022.

\bibitem[Karmakar et~al.(2011)Karmakar, Biswas, Bhowmick, and
  Bhattacharya]{karmakar2011construction}
Karmakar, N., Biswas, A., Bhowmick, P., and Bhattacharya, B.~B.
\newblock Construction of 3d orthogonal cover of a digital object.
\newblock In \emph{Combinatorial Image Analysis: 14th International Workshop,
  IWCIA 2011, Madrid, Spain, May 23-25, 2011. Proceedings 14}, pp.\  70--83.
  Springer, 2011.

\bibitem[Knigge et~al.(2023)Knigge, Romero, Gu, Gavves, Bekkers, Tomczak,
  Hoogendoorn, and jakob Sonke]{knigge2023modelling}
Knigge, D.~M., Romero, D.~W., Gu, A., Gavves, E., Bekkers, E.~J., Tomczak,
  J.~M., Hoogendoorn, M., and jakob Sonke, J.
\newblock Modelling long range dependencies in \$n\$d: From task-specific to a
  general purpose {CNN}.
\newblock In \emph{The Eleventh International Conference on Learning
  Representations}, 2023.
\newblock URL \url{https://openreview.net/forum?id=ZW5aK4yCRqU}.

\bibitem[Le \& Duan(2018)Le and Duan]{le2018pointgrid}
Le, T. and Duan, Y.
\newblock Pointgrid: A deep network for 3d shape understanding.
\newblock In \emph{Proceedings of the IEEE conference on computer vision and
  pattern recognition}, pp.\  9204--9214, 2018.

\bibitem[Liu et~al.(2019)Liu, Tang, Lin, and Han]{liu2019point}
Liu, Z., Tang, H., Lin, Y., and Han, S.
\newblock Point-voxel cnn for efficient 3d deep learning.
\newblock \emph{Advances in Neural Information Processing Systems}, 32, 2019.

\bibitem[Loshchilov \& Hutter(2017)Loshchilov and Hutter]{loshchilov2017sgdr}
Loshchilov, I. and Hutter, F.
\newblock {SGDR}: Stochastic gradient descent with warm restarts.
\newblock In \emph{International Conference on Learning Representations}, 2017.
\newblock URL \url{https://openreview.net/forum?id=Skq89Scxx}.

\bibitem[Loshchilov \& Hutter(2019)Loshchilov and
  Hutter]{loshchilov2018decoupled}
Loshchilov, I. and Hutter, F.
\newblock Decoupled weight decay regularization.
\newblock In \emph{International Conference on Learning Representations}, 2019.
\newblock URL \url{https://openreview.net/forum?id=Bkg6RiCqY7}.

\bibitem[Maturana \& Scherer(2015)Maturana and Scherer]{maturana2015voxnet}
Maturana, D. and Scherer, S.
\newblock Voxnet: A 3d convolutional neural network for real-time object
  recognition.
\newblock In \emph{2015 IEEE/RSJ international conference on intelligent robots
  and systems (IROS)}, pp.\  922--928. IEEE, 2015.

\bibitem[Poli et~al.(2023)Poli, Massaroli, Nguyen, Fu, Dao, Baccus, Bengio,
  Ermon, and R{\'e}]{poli2023hyena}
Poli, M., Massaroli, S., Nguyen, E., Fu, D.~Y., Dao, T., Baccus, S., Bengio,
  Y., Ermon, S., and R{\'e}, C.
\newblock Hyena hierarchy: Towards larger convolutional language models.
\newblock \emph{arXiv preprint arXiv:2302.10866}, 2023.

\bibitem[Qi et~al.(2016)Qi, Su, Nie{\ss}ner, Dai, Yan, and
  Guibas]{qi2016volumetric}
Qi, C.~R., Su, H., Nie{\ss}ner, M., Dai, A., Yan, M., and Guibas, L.~J.
\newblock Volumetric and multi-view cnns for object classification on 3d data.
\newblock In \emph{Proceedings of the IEEE conference on computer vision and
  pattern recognition}, pp.\  5648--5656, 2016.

\bibitem[Qi et~al.(2017{\natexlab{a}})Qi, Su, Mo, and Guibas]{qi2017pointnet}
Qi, C.~R., Su, H., Mo, K., and Guibas, L.~J.
\newblock Pointnet: Deep learning on point sets for 3d classification and
  segmentation.
\newblock In \emph{Proceedings of the IEEE conference on computer vision and
  pattern recognition}, pp.\  652--660, 2017{\natexlab{a}}.

\bibitem[Qi et~al.(2017{\natexlab{b}})Qi, Yi, Su, and Guibas]{qi2017pointnet++}
Qi, C.~R., Yi, L., Su, H., and Guibas, L.~J.
\newblock Pointnet++: Deep hierarchical feature learning on point sets in a
  metric space.
\newblock \emph{Advances in neural information processing systems}, 30,
  2017{\natexlab{b}}.

\bibitem[Ramakrishnan et~al.(2014)Ramakrishnan, Dral, Rupp, and
  Von~Lilienfeld]{ramakrishnan2014quantum}
Ramakrishnan, R., Dral, P.~O., Rupp, M., and Von~Lilienfeld, O.~A.
\newblock Quantum chemistry structures and properties of 134 kilo molecules.
\newblock \emph{Scientific data}, 1\penalty0 (1):\penalty0 1--7, 2014.

\bibitem[Romero \& Lohit(2022)Romero and Lohit]{romero2022learning}
Romero, D.~W. and Lohit, S.
\newblock Learning partial equivariances from data.
\newblock \emph{Advances in Neural Information Processing Systems},
  35:\penalty0 36466--36478, 2022.

\bibitem[Romero et~al.(2021)Romero, Kuzina, Bekkers, Tomczak, and
  Hoogendoorn]{romero2021ckconv}
Romero, D.~W., Kuzina, A., Bekkers, E.~J., Tomczak, J.~M., and Hoogendoorn, M.
\newblock Ckconv: Continuous kernel convolution for sequential data.
\newblock \emph{arXiv preprint arXiv:2102.02611}, 2021.

\bibitem[Ronneberger et~al.(2015)Ronneberger, Fischer, and
  Brox]{ronneberger2015u}
Ronneberger, O., Fischer, P., and Brox, T.
\newblock U-net: Convolutional networks for biomedical image segmentation.
\newblock In \emph{Medical Image Computing and Computer-Assisted
  Intervention--MICCAI 2015: 18th International Conference, Munich, Germany,
  October 5-9, 2015, Proceedings, Part III 18}, pp.\  234--241. Springer, 2015.

\bibitem[Satorras et~al.(2021)Satorras, Hoogeboom, and Welling]{satorras2021n}
Satorras, V.~G., Hoogeboom, E., and Welling, M.
\newblock E (n) equivariant graph neural networks.
\newblock In \emph{International conference on machine learning}, pp.\
  9323--9332. PMLR, 2021.

\bibitem[Sch{\"u}tt et~al.(2017)Sch{\"u}tt, Kindermans, Sauceda~Felix, Chmiela,
  Tkatchenko, and M{\"u}ller]{schutt2017schnet}
Sch{\"u}tt, K., Kindermans, P.-J., Sauceda~Felix, H.~E., Chmiela, S.,
  Tkatchenko, A., and M{\"u}ller, K.-R.
\newblock Schnet: A continuous-filter convolutional neural network for modeling
  quantum interactions.
\newblock \emph{Advances in neural information processing systems}, 30, 2017.

\bibitem[Shi \& Rajkumar(2020)Shi and Rajkumar]{shi2020point}
Shi, W. and Rajkumar, R.
\newblock Point-gnn: Graph neural network for 3d object detection in a point
  cloud.
\newblock In \emph{Proceedings of the IEEE/CVF conference on computer vision
  and pattern recognition}, pp.\  1711--1719, 2020.

\bibitem[Sitzmann et~al.(2020)Sitzmann, Martel, Bergman, Lindell, and
  Wetzstein]{sitzmann2020implicit}
Sitzmann, V., Martel, J., Bergman, A., Lindell, D., and Wetzstein, G.
\newblock Implicit neural representations with periodic activation functions.
\newblock \emph{Advances in Neural Information Processing Systems},
  33:\penalty0 7462--7473, 2020.

\bibitem[Tancik et~al.(2020)Tancik, Srinivasan, Mildenhall, Fridovich-Keil,
  Raghavan, Singhal, Ramamoorthi, Barron, and Ng]{tancik2020fourier}
Tancik, M., Srinivasan, P., Mildenhall, B., Fridovich-Keil, S., Raghavan, N.,
  Singhal, U., Ramamoorthi, R., Barron, J., and Ng, R.
\newblock Fourier features let networks learn high frequency functions in low
  dimensional domains.
\newblock \emph{Advances in Neural Information Processing Systems},
  33:\penalty0 7537--7547, 2020.

\bibitem[Turk \& Levoy(1994)Turk and Levoy]{turk1994zippered}
Turk, G. and Levoy, M.
\newblock Zippered polygon meshes from range images.
\newblock In \emph{Proceedings of the 21st annual conference on Computer
  graphics and interactive techniques}, pp.\  311--318, 1994.

\bibitem[Wang \& Lu(2019)Wang and Lu]{wang2019voxsegnet}
Wang, Z. and Lu, F.
\newblock Voxsegnet: Volumetric cnns for semantic part segmentation of 3d
  shapes.
\newblock \emph{IEEE transactions on visualization and computer graphics},
  26\penalty0 (9):\penalty0 2919--2930, 2019.

\bibitem[Wu et~al.(2019)Wu, Qi, and Fuxin]{wu2019pointconv}
Wu, W., Qi, Z., and Fuxin, L.
\newblock Pointconv: Deep convolutional networks on 3d point clouds.
\newblock In \emph{Proceedings of the IEEE/CVF Conference on computer vision
  and pattern recognition}, pp.\  9621--9630, 2019.

\bibitem[Wu et~al.(2015)Wu, Song, Khosla, Yu, Zhang, Tang, and Xiao]{wu20153d}
Wu, Z., Song, S., Khosla, A., Yu, F., Zhang, L., Tang, X., and Xiao, J.
\newblock 3d shapenets: A deep representation for volumetric shapes.
\newblock In \emph{Proceedings of the IEEE conference on computer vision and
  pattern recognition}, pp.\  1912--1920, 2015.

\bibitem[Xu et~al.(2019)Xu, Lin, Wang, Wang, Cai, Song, Lai, and
  Pei]{xu2019deep}
Xu, Y., Lin, K., Wang, S., Wang, L., Cai, C., Song, C., Lai, L., and Pei, J.
\newblock Deep learning for molecular generation.
\newblock \emph{Future medicinal chemistry}, 11\penalty0 (6):\penalty0
  567--597, 2019.

\bibitem[Yi et~al.(2016)Yi, Kim, Ceylan, Shen, Yan, Su, Lu, Huang, Sheffer, and
  Guibas]{shapenet2016}
Yi, L., Kim, V.~G., Ceylan, D., Shen, I.-C., Yan, M., Su, H., Lu, C., Huang,
  Q., Sheffer, A., and Guibas, L.
\newblock A scalable active framework for region annotation in 3d shape
  collections.
\newblock \emph{ACM Trans. Graph.}, 35\penalty0 (6), dec 2016.
\newblock ISSN 0730-0301.
\newblock \doi{10.1145/2980179.2980238}.
\newblock URL \url{https://doi.org/10.1145/2980179.2980238}.

\bibitem[Zhang et~al.(2022)Zhang, Wan, Shen, and Wu]{zhang2022pvt}
Zhang, C., Wan, H., Shen, X., and Wu, Z.
\newblock Pvt: Point-voxel transformer for point cloud learning.
\newblock \emph{International Journal of Intelligent Systems}, 37\penalty0
  (12):\penalty0 11985--12008, 2022.

\bibitem[Zhao et~al.(2021)Zhao, Jiang, Jia, Torr, and Koltun]{zhao2021point}
Zhao, H., Jiang, L., Jia, J., Torr, P.~H., and Koltun, V.
\newblock Point transformer.
\newblock In \emph{Proceedings of the IEEE/CVF international conference on
  computer vision}, pp.\  16259--16268, 2021.

\end{thebibliography}
\bibliographystyle{icml2023}

\newpage
\appendix
\section*{\Large Appendix}
\section{Network structure and convolution blocks}
\label{app:networkarch}
\begin{figure}[H]
  \centering
  \includegraphics[width=\linewidth]{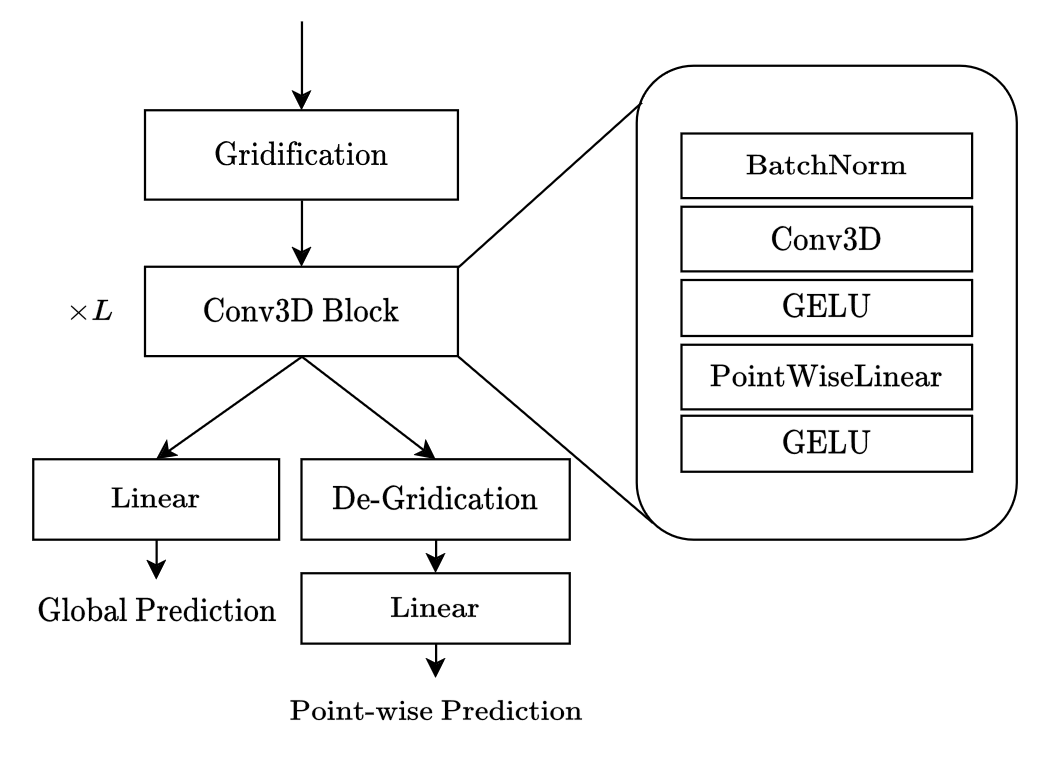}
  \caption{Practical pipeline and convolution blocks.}
  \label{fig:networkarch}
\end{figure}

Fig. \ref{fig:networkarch} shows how we instantiated the grid network as described in section \ref{section:gridnetworks} in practice. The \texttt{Conv3D} blocks are CCNN blocks as in \citet{knigge2023modelling}.

\section{Hyperparameters}
\label{app:hyperparams}
Table \ref{tab:hyperparams} contains the best hyperparameters for the specific datasets found through hyperparameter sweeps.

\begin{table}[H]
\centering
    \begin{minipage}{\textwidth}
    \centering
    \caption{Hyperparameter settings for the different datasets}
    \label{tab:hyperparams}
    \vspace{-2.5mm}
    \begin{small}
    \scalebox{0.85}{
    \begin{tabular}{llll}
    \toprule
     & ModelNet40 & ShapeNet\\
     \midrule
batch size & 32 & 16 \\
nr. conv blocks & 3  &  6  \\
hidden channels & 128  & 256  \\
nr. epochs & 60 &  50  \\
nr. input points & 1000 & 2047 \\ 
$\Omega$ position embedding & 0.1  & 1.0 \\
optimizer & AdamW &  AdamW   \\
learning rate & 0.005  & 0.001  \\
learning rate scheduler & Cosine Annealing &  Cosine Annealing\\
learning rate warmup & 10  &  10 \\
nr. neighbors & 9  & 9 \\
grid resolution & 9 & 13 \\
conv. kernel size &9  & 9\\
dropout & 0.1 & 0.3\\
weight decay & 0 & 0.001\\
aggregation & mean & max \\
    \bottomrule
    \end{tabular}}
    \end{small}
    \end{minipage}%
    \vspace{-4mm}
\end{table}

\end{document}